\documentclass[conference]{IEEEtran}
\IEEEoverridecommandlockouts
\usepackage{cite}
\usepackage{amsmath,amssymb,amsfonts}
\usepackage{algorithmic}
\usepackage{graphicx}
\usepackage{textcomp}
\usepackage{xcolor}
\def\BibTeX{{\rm B\kern-.05em{\sc i\kern-.025em b}\kern-.08em
		T\kern-.1667em\lower.7ex\hbox{E}\kern-.125emX}}
\begin{document}
	
	\title{Correlation Analysis of Generative Models\\
		{\footnotesize \textsuperscript{}}
	}
	
	\author{\IEEEauthorblockN{Zhengguo Li}
		\IEEEauthorblockA{VI Department\\Institute for Infocomm Research\\ A*STAR,	Singapore, 138632 \\
			Li\_Zhengguo@a-star.edu.sg}
		\and
			\IEEEauthorblockN{Chaobing Zheng}
			\IEEEauthorblockA{School of ISE\\ 
			Wuhan University of \\ Science and Technology\\
			Wuhan, China \\
			zhengchaobing@wust.edu.cn}
		\and
		\IEEEauthorblockN{Wei Wang}
		\IEEEauthorblockA{School of CST\\
			Wuhan University of \\ Science and Technology\\
			Wuhan, China \\
			wangwei8@wust.edu.cn}
	}
	
\maketitle

\begin{abstract}
Based on literature review about existing diffusion models and flow matching with a neural network to predict a predefined target from noisy data, a unified representation is first proposed for these models using two simple linear  equations in this paper.  Theoretical analysis of the proposed model is then presented. Our theoretical analysis shows that the correlation between the noisy data and the predicted target is sometimes weak in the existing diffusion models and flow matching. This might affect the prediction (or learning) process  which plays a crucial role in all models.
\end{abstract}

\section{Introduction}
The goal of generative model is to use a set of data which is sampled from an unknown distribution $p_{d}(\cdot)$ to learn a model for generating new samples from the distribution $p_{d}(\cdot)$. Diffusion models and flow matching have emerged as promising and popular frameworks for generative models \cite{1sohl2015,1ho2020,1Salimans2021,1liu2022,1lu2024,1chentq2018,1lipman2023,1roy2024,1black2024}, demonstrating state-of-the-art performance across many applications \cite{1rombach2022,1gu2022,1liuxc2024,1shao2025,1fang2025,1liang2025,1chi2025,1joha2025,1lv2025,1wang2025}. Among these applications, the depth estimation in \cite{1shao2025}, the diffusion planning in \cite{1liang2025}, the diffusion policies in \cite{1chi2025,1lv2025}, the vision-language-action flow model in \cite{1black2024}, and the FlowRAM \cite{1wang2025} are highly demanded for robotics and embodied intelligence.

Each diffusion model has a forward (or diffusion) process and a reverse (or generative) process. The forward (or diffusion) process begins with a ground truth data sample and evolves to a noise sample by progressively adding the Gaussian noise to the data sample. This paper focuses on the broad class of models where a neural network is trained to map the noisy data to a predicted target during the  forward (or diffusion) process. The output of the neural network and the noisy data will be utilized to estimate the ground-truth data and the Gaussian noise in the reverse (or generative) process. The reverse (or generative) process starts from the noise sample and converts it back to the corresponding data sample through progressively reducing the Gaussian noise from the noisy data. The reverse (or generative) process is conducted via tens to thousands steps. Clearly, these models face a drawback in terms of slow sampling speeds, requiring a large number of network (or model) evaluations. Thus, it is desired to optimize the sampling process using trajectory distillation to enhance efficiency without compromising sample quality  \cite{1Salimans2021}.

The fitting error of the neural network could be amplified during the trajectory distillation \cite{1Salimans2021}. The Gaussian noise is the target to be predicted by the neural network in \cite{1ho2020} and the noise prediction network in \cite{1ho2020} might not be well suited for the distillation. When the network is evaluated at  signal-to-noise ratios around zeros, the fitting error of the neural network is amplified significantly when the output of the neural network is adopted to implicitly predict the data. Since later updates can correct the error, this is not a problem if the reverse process has many steps. However, it becomes an issue for the trajectory distillation with only a few steps. The issue was addressed in \cite{1Salimans2021,1gu2022} by choosing the ground-truth data as the target to be predicted. The common framework in \cite{1Karras2022} and consistency model in \cite{1song2023} as well as flow matching \cite{1liu2022,1lu2024} were also proposed to address this issue by setting the target to be predicted as a combination of the Gaussian noise and ground-truth data. As such, the amplification of the fitting error is well addressed. It is indeed important to reduce  the amplification of the fitting error in the reverse (or generative) process. One natural question to think about is ``Is there any other issue that is ignored by all existing models?".

The objective of this paper is to answer this question. A unified representation is first presented using two simple linear equations. With strong support from mathematical derivations, the variances of the ground-truth data and the noise are selected as identity matrices, and the neural network is designed to map the noisy data and time instant to the predicted target in the unified model. The proposed model can be adopted to represent all existing models \cite{1ho2020,1rombach2022,1Salimans2021,1gu2022,1Karras2022,1song2023,1liu2022,1lu2024}. Interestingly, the reverse (or generative) processes of all models can be easily obtained by solving the linear equations. In addition to analyzing the amplification factors of the fitting error, the Pearson correlation between the noisy data and the predicted target is also studied for all models.  Our investigation indicates that  the correlation between the noisy data and the predicted target is ignored by all existing models. The weak correlation between the noisy data and the predicted target might be a concern of existing models. The weak correlation might affect the performance of the neural network which plays an important role in all models. To the best of our knowledge, we are the first to study the correlation between the noisy data and the predicted target for the diffusion models and flow matching. Our investigation provide a new insight about the diffusion models and flow matching. It should be mentioned that the contribution of this paper is to provide theoretical analysis of the diffusion models and flow matching rather than experiment results on any particular diffusion model or flow matching.

The rest of this paper is organized as below. Typical  diffusion models and flow matching are presented and analyzed in Section \ref{existing}. A unified model is proposed in Section \ref{unified} using slime linear time-varying equations. Theoretical analysis of the existing models is then provided in Section \ref{analysis}. Finally, conclusion remarks are given in Section \ref{conclusion}.

\section{Typical Models}
\label{existing}
In this section, typical diffusion models and flow matching are presented and analyzed with emphasis on 1) the standard deviation of the ground truth data, 2) the variance of the noise, 3) the choice of the predicted target, 4) inputs of the neural network, and 5) noise-adding scheme in the forward (or diffusion) process. 
\subsection{Diffusion Model}

The  diffusion model consists of a forward (or diffusion) process and a reverse (or generative) process \cite{1ho2020,1rombach2022,1Salimans2021,1gu2022}. Let $Z(\sim p_{d}(Z))$ be training data from an unknown data distribution $p_{d}(Z)$ with the standard deviation as one, and $\mathcal{N}(\cdot)$ be a Gaussian distribution. $Z$ can come from a latent space \cite{1rombach2022}.
$X=\{X_t|t\in [t_0, t_f]$ be the latent variables of a diffusion model \cite{1ho2020}.
The forward process  is specified by a noise schedule consisting of  differentiable functions $\alpha_t$ and $\sigma_t$
such that the log signal-to-noise-ratio (i.e., $\lambda_t = \log[\alpha_t^2 /\sigma_t^2]$) decreases monotonically with $t$.
The forward process generates a sequence of noisy variables  by gradually adding Gaussian noise to  $Z$  as \cite{1sohl2015}
\begin{align}
	\label{DDPM1}
	X_t=\alpha_tZ+\sigma_t\epsilon,
\end{align}
where $\epsilon(\sim \mathcal{N}(0,I))$ is the standard Gaussian noise. $\alpha_t\in(0,1)$ represents the noise schedule, $\alpha_{t_0}\rightarrow 1$ and $\alpha_{t_f}\rightarrow 0$.
$\sigma_t$ controls the noise variance. Since 
\begin{align}
\label{alphasigma}
\alpha_t^2+\sigma_t^2=1,  
\end{align}
the diffusion process is a standard variance preserving process.

The reverse process is to iteratively remove noise and convert the noise sample $X_{t_f}(\sim \mathcal{N}(\alpha_{t_f} Z, \sigma_{t_f}^2 I))$ back to
the corresponding data sample $Z$ step by step using \cite{1song2021}
\begin{align}
	\label{qxt-1}
	q(X_{t'}|X_t, Z)=\mathcal{N}(X_{t'};\mu_t(X_t,Z),\varsigma_t^2I)\; ;\; t'<t,
\end{align}
where $\mu_t(X_t,Z)$ is
\begin{align}
	\label{mutxtz}
	\mu_t(X_t,Z)=\alpha_{t'}Z+\sqrt{1-\alpha_{t'}^2-\varsigma_t^2}\frac{X_t-\alpha_t Z}{\sqrt{1-\alpha_t^2}}.
\end{align}

For simplicity, let $\beta_t$ be $(1-\frac{\alpha_t^2}{\alpha_{t'}^2})$. By choosing $\varsigma_t^2$ as
\begin{align}
	\varsigma_t^2=\frac{1-\alpha_{t'}^2}{1-\alpha_t^2}\beta_t,
\end{align}
$q(X_{t'}|X_t, Z)$ in equation (\ref{qxt-1}) becomes the reverse  process of the statistic denoising diffusion probabilistic model (DDPM) in \cite{1ho2020}.

The deterministic  denoising diffusion implicit model (DDIM) in \cite{1song2021} requires that $\varsigma_t$ is zero except for $t$ as $t_0$. The deterministic rule is adopted in this paper. The corresponding reverse process is
\begin{align}
	X_{t'}=\alpha_{t'}Z+\sqrt{1-\alpha_{t'}^2}\frac{X_t-\alpha_t Z}{\sqrt{1-\alpha_t^2}}.
\end{align}

It is required by the reverse process to estimate the ground-truth $Z$ from the noisy data $X_t$. Let $\omega$ be a target to be predicted or learned by a neural network $f_{\theta}(X_t,t)$. Two popular choices are given as follows:

1) $\omega$ is selected as the Gaussian noise $\epsilon$, and it is a noise prediction network \cite{1ho2020,1rombach2022}. Once the Gaussian noise $\epsilon$ is predicted, the ground-truth $Z$ is estimated from
equation (\ref{DDPM1}) as
\begin{align}
Z=\frac{1}{\alpha_t}X_t-\frac{\sigma_t}{\alpha_t}f_{\theta}(X_t,t).
\end{align}

2) $\omega$ is selected as the data $Z$, and it is a data prediction network \cite{1Salimans2021,1gu2022}. Once the  ground-truth $Z$ is  predicted, the noise $\epsilon$ is estimated from
equation (\ref{DDPM1}) as
\begin{align}
\epsilon=\frac{1}{\sigma_t}X_t-\frac{\alpha_t}{\sigma_t}f_{\theta}(X_t,t).
\end{align}

It should be pointed out that the neural network $f_{\theta}(X_t,t)$ is trained in the forward (or diffusion) process rather than the reverse (or generative) process.

\subsection{Common Framework}

The forward process of the common framework in \cite{1Karras2022} is represented by
\begin{align}
	\label{cm1}
	Y_t=\tilde{Z}+\tilde{\epsilon}_t, 
\end{align}
where the standard deviation  of $\tilde{Z}$ is $\sigma_d$, and $\epsilon_t\sim \mathcal{N}(0,t^2I)$.

Defining a set of parameters as \cite{1Karras2022}
\begin{align}
\label{cm2}
	\left\{\begin{array}{l}
		c_{in}(t)=\frac{1}{\sqrt{\sigma_d^2+t^2}},\\
		c_{out}(t)=\frac{\sigma_dt}{\sqrt{\sigma_d^2+t^2}}\\
		c_{skip}(t)=\frac{\sigma_d^2}{\sigma_d^2+t^2}
	\end{array}
	\right.,
\end{align}
a neural network $f_{\theta}(c_{in}(t)Y_t,c_{noise}(t))$ is trained by the common framework in \cite{1Karras2022}, it maps $c_{in}(t)Y_t$ and $c_{noise}(t)$ to the following target:
\begin{align}
	\label{cm3}
	\omega=\frac{1}{c_{out(t)}}(\tilde{Z}-c_{skip}(t)Y_t).
\end{align}
The consistency model  \cite{1song2023} is also on top of equations (\ref{cm1})-(\ref{cm3}). To prevent the time derivative $\frac{\partial f_{\theta}}{\partial t}$ from being amplified and the numerical instability, $c_{noise}(t)$ is selected as $t$ in \cite{1lu2024,1chenjs2025}.

Introducing a set of transformations as
\begin{align}
\left\{\begin{array}{l}
	X_t=c_{in}(t)Y_t\\
    Z=\frac{\tilde{Z}}{\sigma_d}\\
    \epsilon=\frac{\epsilon_t}{t}
    \end{array}
    \right.,
\end{align}
and defining $\alpha_t$ and $\sigma_t$ as
\begin{align}
\label{sigmat}
\left\{\begin{array}{l}
	\alpha_t=\frac{\sigma_d}{\sqrt{\sigma_d^2+t^2}}\\
	\sigma_t=\frac{t}{\sqrt{\sigma_d^2+t^2}}
    \end{array}
    \right.,
\end{align}
equation (\ref{DDPM1}) can be derived from equations (\ref{cm1}) and (\ref{cm2}).  It can be easily verified that equation (\ref{alphasigma}) holds. 
Thus, the common network \cite{1Karras2022} and the consistency model \cite{1song2023} are also variance preserving processes.   Furthermore, 
it follows from equations (\ref{cm1})-(\ref{sigmat}) that the predicted target is represented by 
\begin{align}
	\omega = \sigma_t Z-\alpha_t \epsilon. 
\end{align}

From equations (\ref{cm1})-(\ref{sigmat}), it is reasonable to assume that {\it the standard deviations of the ground truth data $Z$ and the noise $\epsilon$ are ones, and inputs of the neural network $f_{\theta}(\cdot)$ are fixed as the noisy data $X_t$ and time instant $t$ for all models}.

\subsection{Flow Matching}

$Z\sim \pi_0$ and $\epsilon\sim \pi_1$ are two empirical observations. The rectified flow induced from
$(Z,\epsilon)$ is an ordinary differentiable equation (ODE) with $t_0$ and $t_f$ as 0 and 1 \cite{1liu2022}. The forward (or diffusion) process is represented by a linear combination of ground-truth data $Z$ and Gaussian noise $\epsilon$ as 
\begin{align}
	\label{rectifiedflow}
	X_t = (1-t)Z+t\epsilon,
\end{align}
which converts $Z$ from $\pi_0$ to $\epsilon$ following $\pi_1$. Without loss of generality, the standard deviations of $Z$ and $\epsilon$ are one's.

Since $((1-t)^2+t^2)$ is not equal to 1, the rectified flow (\ref{rectifiedflow}) is not a variance preserving process. The target to be predicted or learned by neural network $f_{\theta}(X_t,t)$ is defined as
\begin{align}
	\omega=\frac{d X_t}{dt}=\epsilon-Z.
\end{align}

Flow matching theoretically learns a constant vector field to achieve a straight flow from the Gaussian noise $\epsilon$ to the ground-truth data $Z$.  Few-step or even single-step sampling is enabled in an ideal situation. However, it is very challenging for the flow matching to learn the constant velocity field during actual training. The resulting trajectories still remain curved. The curvature is arguably attributed to the crossing of reference trajectories during training and  a reflow is employed  to mitigate this issue. As shown in \cite{1daijm2025}, the performance of the reflow can be improved.  

The Trigflow \cite{1lu2024} is inspired by the common framework and the rectified flow. The forward (or diffusion) process is
modeled by
\begin{align}
	X_t=\cos(t)Z+\sin(t)\epsilon\; ;\; t\in[0,\frac{\pi}{2}],
\end{align}
and the target to be predicted or learned by neural network $f_{\theta}(X_t,t)$ is defined as
\begin{align}
	\omega=\frac{d X_t}{dt}=-\sin(t)Z+\cos(t)\epsilon.
\end{align}

The TrigFlow is a  variance preserving process, and it unifies the common framework \cite{1Karras2022,1song2023} and flow matching \cite{1liu2022,1lu2024}. With the Triflow, the diffusion process, diffusion model parameterization, diffusion training objective, and consistency model parameterization have simple expressions.

The forward process of the Trigflow  can be replaced by
\begin{align}
	X_t=\cos(\frac{\pi}{2}t)Z+\sin(\frac{\pi}{2}t)\epsilon\; ;\; t\in[0,1],
\end{align}
and the target to be predicted can be replaced by 
\begin{align}
	\omega=-\sin(\frac{\pi}{2}t)Z+\cos(\frac{\pi}{2}t)\epsilon.
\end{align}

\begin{table*}[htb]
	\begin{center}
		\begin{tabular}{c|c|c|c|c|c}  \hline
			& diffusion model in \cite{1ho2020,1rombach2022}  & diffusion model in \cite{1Salimans2021,1gu2022} & models in \cite{1Karras2022,1song2023} & Trigflow in \cite{1lu2024}  & flow matching in \cite{1liu2022} \\\hline
			$a_{11}(t)$  & $\alpha_t$ & $\alpha_t$  & $\frac{\sigma_d}{\sqrt{\sigma_d^2+t^2}}$ &$\cos(t)$ & $1-t$ \\\hline
			$a_{12}(t)$ & $\sigma_t$ & $\sigma_t$ & $\frac{t}{\sqrt{\sigma_d^2+t^2}}$
			& $\sin(t)$  & $t$ \\\hline
			$a_{21}(t)$  & $0$ & $1$ & $a_{12}(t)$ & $-a_{12}(t)$ & $-1$\\\hline
			$a_{22}(t)$ & $1$ & $0$ &  $-a_{11}(t)$ & $a_{11}(t)$&$1$\\\hline
			$|A(t)|$ & $\alpha_t$ & $-\sigma_t$ & $-1$ & 1 &1\\\hline
			$\Psi_{X_t,\omega}$ & $\sigma_t$& $\alpha_t$& 0 & 0& $\frac{2t-1}{\sqrt{2((t-1)^2+t^2)}}$\\\hline
			$|\Phi(t,t')|$ & $\frac{|\sigma_{t'}\alpha_t-\sigma_t\alpha_{t'}|}{\alpha(t)}$& $\frac{|\sigma_{t'}\alpha_t-\sigma_t\alpha_{t'}|}{\sigma_t}$&
			$\frac{\sigma_d(t-t')}{\sqrt{\sigma_d^2+t^2}\sqrt{\sigma_d^2+(t')^2}}$& $\sin(t-t')$& $t-t'$ \\\hline
            			$|\Phi(t,0)|$ & $\frac{\sigma_t}{\alpha(t)}$& 1&
			$\frac{t}{\sqrt{\sigma_d^2+t^2}}$& $\sin(t)$& $t$ \\\hline
		\end{tabular}
	\end{center}
	\caption{Theoretical analysis of existing diffusion models and flow matching.  For simplicity, the deterministic method in \cite{1song2021} is chosen for illustration.}
	\label{theoreticalanalysis}
\end{table*}

\section{A Unified Representation}
\label{unified}

In this section, a unified representation is first proposed for
the forward (or diffusion) process and prediction (or learning) process of the existing diffusion models \cite{1ho2020,1rombach2022,1Salimans2021,1gu2022,1Karras2022,1song2023} and flow matching \cite{1liu2022,1lu2024}.

 Let $Z(\sim p_{d}(Z))$ be training data from an unknown data distribution $p_{d}(Z)$, $\epsilon$ be the Gaussian noise, $\omega$ be the predicted target, and $f_{\theta}(\cdot)$ be the neural network for the prediction of the target $\omega$. According to the mathematical derivations of the common framework in the previous section, the standard deviation of the ground-truth data is unified as 1, the noise $\epsilon$ is unified as $\epsilon\sim\mathcal{N}(0,I)$. The neural network is unified as $f_{\theta}(X_t,t)$. The forward (or diffusion) process is from $t_0$ to $t_f$.  The forward (or diffusion) process and prediction (or learning) process
of the existing  models  are unified as
\begin{align}
	\nonumber
	\left[\begin{array}{l}
		X_t\\
		f_{\theta}(X_t,t)\\
	\end{array}
	\right]=&A(t)\left[\begin{array}{l}
		Z\\
		\epsilon\\
	\end{array}
	\right]\\\label{linearmodel}
	=&\left[\begin{array}{l l}
		a_{11}(t) & a_{12}(t)\\
		a_{21}(t) & a_{22}(t)\\
	\end{array}
	\right]\left[\begin{array}{l}
		Z\\
		\epsilon\\
	\end{array}
	\right],
\end{align}
where  $a_{11}(t)$, $a_{12}(t)$, $a_{21}(t)$ and $a_{22}(t)$ are given for each model in Table \ref{theoreticalanalysis}. 

Let $|A(t)|$ be the determinant of matrix $A(t)$. $|A(t)|$ is also provided for each model in Table \ref{theoreticalanalysis}. Clearly, the matrix $A(t)$ has full rank except at time $t_0$ for the diffusion model \cite{1Salimans2021,1gu2022} or at time $t_f$ for the diffusion model \cite{1ho2020,1rombach2022}. It can be easily verified that
\begin{align}
	\left\{\begin{array}{l}
		\lim_{t\rightarrow t_0}a_{11}(t)=1\\
		\lim_{t\rightarrow t_f}a_{11}(t)=0\\
		\lim_{t\rightarrow t_0}a_{12}(t)=0\\
		\lim_{t\rightarrow t_f}a_{12}(t)=1
	\end{array}
	\right.
\end{align}
hold for all models \cite{1ho2020,1rombach2022,1Salimans2021,1gu2022,1Karras2022,1song2023,1liu2022,1lu2024} and
\begin{align}
	\left\{\begin{array}{l}
		\frac{d a_{11}(t)}{dt}=a_{21}(t)\\
		\frac{d a_{12}(t)}{dt}=a_{22}(t)
	\end{array}
	\right.
\end{align}
hold for the flow matching \cite{1liu2022,1lu2024}.

The reverse (or generative) process is from $t_f$ to $t_0$. Considering two instances $t$ and $t'$ satisfying $t'<t$, the noisy data $X_t$ and the prediction $f_{\theta}(X_t,t)$
are available at $t'$ in the reverse (or generative) process of all models \cite{1ho2020,1rombach2022,1Salimans2021,1gu2022,1Karras2022,1song2023,1liu2022,1lu2024}.
$Z$ and $\epsilon$ can be easily obtained by solving the linear time-varying equation (\ref{linearmodel}) as
\begin{align}
	\label{predictionz}
	\left[\begin{array}{l}
		Z\\
		\epsilon\\
	\end{array}
	\right]=&
	\frac{1}{|A(t)|}\left[\begin{array}{l l}
		a_{22}(t) & -a_{12}(t)\\
		-a_{21}(t) & a_{11}(t)\\
	\end{array}
	\right]\left[\begin{array}{l}
		X_t\\
		f_{\theta}\\
	\end{array}
	\right].
\end{align}
The unified reverse (or generative) process is then given as
\begin{align}
	\label{reverseprocess}
	X_{t'}=&a_{11}(t')Z+a_{12}(t')\epsilon.
\end{align}

A well known drawback of the existing models is their slow sampling speed, often requiring a large number of
steps to generate a single sample. Fortunately, consistency models including \cite{1song2023} provide  an efficient way to address this issue. A consistency model \cite{1song2023} $F_{\theta}(X_t, t)$ maps the noisy data $X_t$ directly to the ground-truth data $Z$ in one step. From equation (\ref{predictionz}), the unified consistency model is given as
\begin{align}
\label{consistencymodel}
	F_{\theta}(X_t,t)=\frac{a_{22}(t)}{|A(t)|}X_t-\frac{a_{12}(t)}{|A(t)|}f_{\theta}(X_t,t).
\end{align}
The consistency model $F_{\theta}(X_t,t)$ is actually the data prediction component of the unified model. Similarly, the unified noise prediction component is provided as
\begin{align}
\label{noiseprediction}
	G_{\theta}(X_t,t)=-\frac{a_{21}(t)}{|A(t)|}X_t+\frac{a_{11}(t)}{|A(t)|}f_{\theta}(X_t,t).
\end{align}

The stochastic differential equation can be  derived from equation (\ref{linearmodel}) as 
\begin{align}
\nonumber
\dot{X}_t&=\dot{a}_{11}(t)Z+\dot{a}_{12}(t)\epsilon\\\nonumber
&=\frac{\dot{a}_{11}(t)}{a_{11}(t)}X_t-a_{12}(t)(\frac{\dot{a}_{11}(t)}{a_{11}(t)}-\frac{\dot{a}_{12}(t)}{a_{12}(t)})\epsilon\\\label{dotxt}
&=\frac{\dot{a}_{11}(t)}{a_{11}(t)}X_t-\frac{a_{12}(t)\dot{\lambda}(t)}{2}\epsilon,
\end{align}
where $\lambda(t)$ $(=\log(a_{11}^2(t)/a_{12}^2(t)))$is the signal-to-noise-ratio. The corresponding flow matching is then derived from equations  (\ref{noiseprediction}) and (\ref{dotxt}) as 
\begin{align}
\nonumber
V_{\theta}(X_t,t)\dot{=}&\dot{X}_t\\
=&\frac{\dot{a}_{11}(t)}{a_{11}(t)}X(t)-\frac{a_{12}(t)\dot{\lambda}(t)}{2}G_{\theta}(X_t,t)\\\nonumber
=&(\frac{\dot{a}_{11}(t)}{a_{11}(t)}+\frac{a_{21}(t)a_{12}(t)\dot{\lambda}(t)}{2|A(t)|})X(t)\end{align}\begin{align}
&-\frac{a_{11}(t)a_{12}(t)\dot{\lambda}(t)}{2|A(t)|}f_{\theta}(X_t,t).
\end{align}

\section{Theoretical Analysis}
\label{analysis}

In this section, the amplification factor of the fitting arrow is first analyzed for the reverse process (\ref{reverseprocess}). 

It can be derived from equations
(\ref{reverseprocess})-(\ref{noiseprediction} ) that
\begin{align}
	\nonumber
	X_{t'}=&a_{11}(t')F_{\theta}(X_t,t)+a_{12}(t')G_{\theta}(X_t,t)\\\nonumber
    =&\frac{a_{11}(t')a_{22}(t)-a_{12}(t')a_{21}(t)}{|A(t)|}X_t\\
	&+\frac{a_{11}(t)a_{12}(t')-a_{11}(t')a_{12}(t)}{|A(t)|}f_{\theta}(X_t,t).
\end{align}

Let the amplification factor of the fitting error be denoted as $\Phi(t,t')$. It is known from the above equation that 
\begin{align}
	\Phi(t,t')=\frac{a_{11}(t)a_{12}(t')-a_{11}(t')a_{12}(t)}{|A(t)|}.
\end{align}

When the reverse (or generative) process has many steps, $t'$ is almost equal to $t$ and $\Phi(t,t')$ is not large. However, this is not true for the trajectory diffusion with a few steps and the consistency model with one step. The fitting error of $f_{\theta}(X_t,t)$ is amplified by  $|\frac{a_{12}(t)}{|A(t)|}|$ in the consistency model (\ref{consistencymodel}). It  can be verified that \begin{align}
 |\Phi(t,0)|=|\frac{a_{12}(t)}{|A(t)|}|.
\end{align}

Since $(a_{11}(t)a_{12}(t')-a_{11}(t')a_{12}(t))$  and $a_{12}(t)$ are independent of the target $\omega$, $|\Phi(t,t')|$ and $|\Phi(t,0)|$ are minimized if and only if $|A(t)|$ is maximized. A popular strategy in \cite{1Karras2022,1song2023,1liu2022,1lu2024} is to choose the target $\omega$ as a  time-varying linear combination of the ground truth data $Z$ and the noise $\epsilon$ such that the determinant $|A(t)|$ is independent of $t$ as demonstrated in Table \ref{theoreticalanalysis}. These choices are attractive because the fitting error of $f_{\theta}(X_t,t)$ can be prevented from being significantly amplified for any  $t$ and any $t'(<t)$. This is highly demanded for the diffusion models and flow matching.

The correlation between the noisy data $X_t$ and the predicted target $\omega$ is now studied. The objective is to provide a new insight on the diffusion models and flow matching which might be useful for future works on them. 

The Pearson correlation has been widely used in many applications, for example,  instance normalization \cite{1ulyanov2016}, non-stationary time series forecasting \cite{1kimts2022}, exposure interpolation for high dynamic range imaging \cite{1lizg2025},  and so on. The Pearson correlation between the noisy data $X_t$ and the predicted target $\omega$ (or $f_{\theta}(X_t,t)$) is defined as
\begin{align}
	\Psi_{X_t,\omega}=\frac{\mbox{cov}(X_t,\omega)}{\sigma_{X_t}\sigma_{\omega}},
\end{align}
where $\mbox{cov}(X_t,\omega)$ is the covariance of the noisy data $X_t$ and the predicted target $\omega$. It follows from equation (\ref{linearmodel}) that
\begin{align}
	\Psi_{X_t,\omega}=\frac{a_{11}(t)a_{21}(t)+a_{12}(t)a_{22}(t)}{\sqrt{a_{11}^2(t)+a_{12}^2(t)}\sqrt{a_{21}^2(t)+a_{22}^2(t)}}.
    \end{align}

The Pearson correlation coefficients of all models \cite{1ho2020,1rombach2022,1Salimans2021,1gu2022,1Karras2022,1song2023,1liu2022,1lu2024}
are given in Table \ref{theoreticalanalysis}. Surprisingly, the Pearson correlation coefficients are zeros for the
models in \cite{1Karras2022,1song2023,1lu2024}. Overall, existing models \cite{1Karras2022,1song2023,1liu2022,1lu2024} focus on minimizing the amplification factors $|\Phi(t,t')|$ and $|\Phi(t,0)|$ while ignoring the correlation between the noisy data $X_t$ and the predicted target $\omega$. However, when the correlation between them is weak, it could be challenging for the neural network $f_{\theta}(X_t,t)$ to predict (or learn) the target $\omega$ from the noisy data $X_t$\footnote{We happened to find the reference \cite{1daijm2025} after we finished this paper. Coincidentally, it was pointed out in \cite{1daijm2025} that ``the neural network model is not an exact solver, intermediate states along different trajectories may become ``approximately crossing" due to their high similarity in statistical property". We believe that this is at least partially due to the weak correlation between the noisy data $X_t$ and the predicted target $\omega$.}. This might be a concern of existing  models \cite{1ho2020,1rombach2022,1Salimans2021,1gu2022,1Karras2022,1song2023,1liu2022,1lu2024}.  Actually, it is already realized in \cite{1esser2024} that the prediction of the target $\omega$ is more difficult for the flow matching when $t$ is  in the middle of $[0,1]$. This is because that $\Psi_{X_t,\omega}$ is almost zero when $t$ is around 0.5. Instead of sampling $t$ using a uniform distribution $U(0, 1)$, Esser et al. \cite{1esser2024}
proposed sampling from a logit-normal distribution that emphasizes the middle time-steps. An alternative solution is to develop new diffusion models and flow matching under the following two requirements:
1) the fitting error of $f_{\theta}(X_t,t)$ is avoided from being significantly amplified for any $t$ and any $t'(<t)$; and 2) the correlation between the noisy data $X_t$ and the predicted target $\omega$ is strong. The alternative solution could be more efficient.

Straighten viscous rectified flow via noise optimization (VRFNO) in \cite{1daijm2025} could be an example in this direction. It can be easily shown that the reparameterization technique in the VERFNO maximizes the Pearson correlation between the ground-truth data $Z$ and the re-parametrized noise $\epsilon$. More alternative solutions would be expected. For example, another reparameterization technique can be developed using the normalization method in \cite{1lizg2025}. The window sizes for the standard deviation computation  and the mean computation are different in \cite{1lizg2025}, while they are the same in \cite{1daijm2025}.

\section{Conclusion Remarks}
\label{conclusion}
A unified representation is proposed in this paper for existing diffusion models and flow matching with a neural network to predict a predefined target from noisy data. Theoretical analysis shows that the correlation between noisy data and predicted target is weak in existing diffusion models and flow marching. The weak correlation might affect the performance of the neural network. Solutions will be provided to address this concern and the solutions will be adopted to study vision-language-action flow model in \cite{1black2024}, embodied intelligence \cite{1chi2025,1lv2025},  meta-lens imaging \cite{1fangf2026}, exposure interpolation \cite{1lizg2025}, physics-informed diffusion model \cite{1bastek2025}, and physics-guided diffusion model    in our future R\&D.

\bibliographystyle{IEEEtran}
\bibliography{refs}

\end{document}